\documentclass{article}

\usepackage{microtype}
\usepackage{graphicx}
\usepackage{subfigure}
\usepackage{booktabs} % for professional tables

\usepackage{hyperref}

\usepackage[accepted]{arxiv}

\usepackage{amsmath}
\usepackage{amssymb}
\usepackage{mathtools}
\usepackage{amsthm}

\usepackage[capitalize,noabbrev]{cleveref}

\usepackage{bbm}
\usepackage{multirow}
\usepackage{graphicx}
\usepackage{amsmath,amsfonts,bm}

\def\eqref#1{equation~\ref{#1}}
\def\1{\bm{1}}

\def\vzero{{\bm{0}}}

\def\vc{{\bm{c}}}

\def\vf{{\bm{f}}}

\def\vh{{\bm{h}}}

\def\vv{{\bm{v}}}
\def\vw{{\bm{w}}}

\DeclareMathAlphabet{\mathsfit}{\encodingdefault}{\sfdefault}{m}{sl}
\SetMathAlphabet{\mathsfit}{bold}{\encodingdefault}{\sfdefault}{bx}{n}

\newcommand{\E}{\mathbb{E}}

\usepackage{amssymb}
\usepackage{pifont}
\newcommand{\eg}{\textit{e}.\textit{g}.}
\newcommand{\logo}{\raisebox{-10.5pt}{\includegraphics[height=30pt]{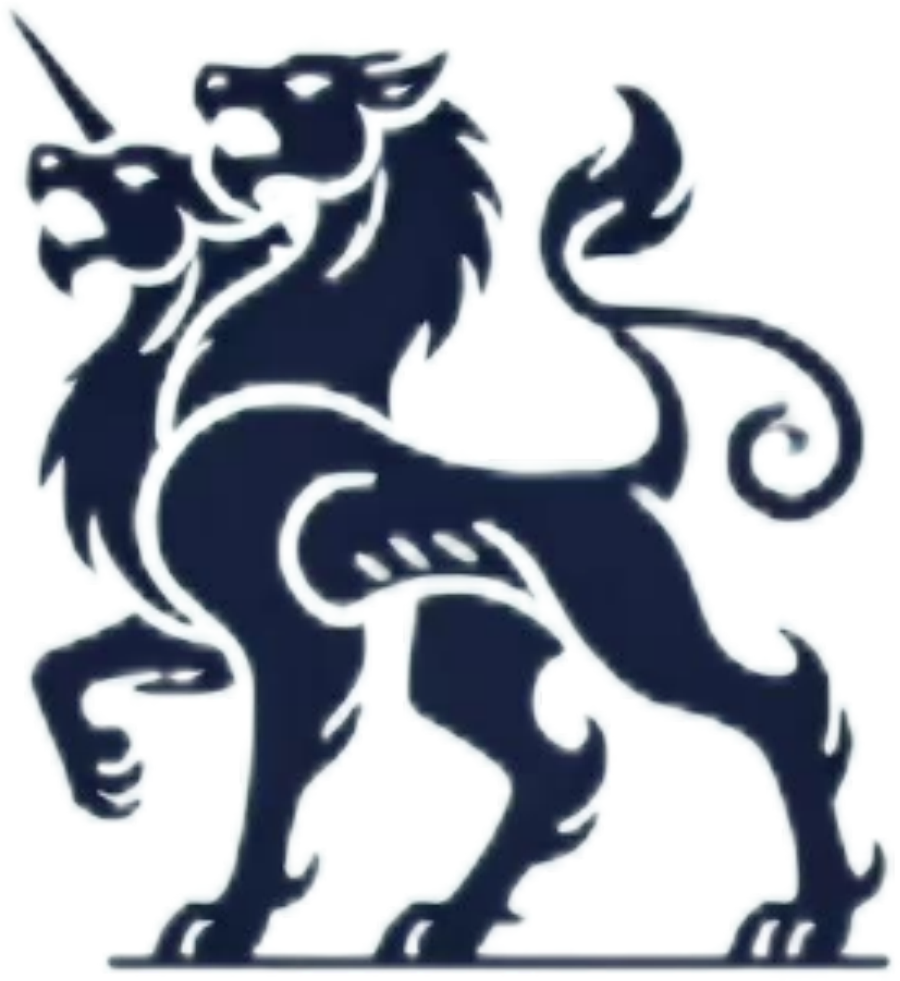}}}

\theoremstyle{plain}

\theoremstyle{definition}

\theoremstyle{remark}

\usepackage[textsize=tiny]{todonotes}

\icmltitlerunning{Orthus: Autoregressive Interleaved Image-Text Generation with Modality-Specific Heads}

\begin{document}

\twocolumn[
\icmltitle{\logo Orthus: Autoregressive Interleaved Image-Text Generation with Modality-Specific Heads}



\icmlsetsymbol{equal}{*}

\begin{icmlauthorlist}
\icmlauthor{Siqi Kou}{sjtu,equal}
\icmlauthor{Jiachun Jin}{sjtu}
\icmlauthor{Zhihong Liu}{sjtu}
\icmlauthor{Chang Liu}{sjtu,equal}
\\
\icmlauthor{Ye Ma}{kuaishou}
\icmlauthor{Jian Jia}{kuaishou}
\icmlauthor{Quan Chen}{kuaishou}
\icmlauthor{Peng Jiang}{kuaishou}
\icmlauthor{Zhijie Deng}{sjtu}
\end{icmlauthorlist}

\icmlaffiliation{sjtu}{Qing Yuan Research Institute, SEIEE, Shanghai Jiao Tong University}
\icmlaffiliation{kuaishou}{Kuaishou Technology}

\icmlcorrespondingauthor{Zhijie Deng}{zhijied@sjtu.edu.cn}


\vskip 0.3in
]



\printAffiliationsAndNotice{\icmlEqualContribution} % otherwise use the standard text.

\begin{abstract}
We introduce Orthus, a unified multimodal model that excels in generating interleaved images and text from mixed-modality inputs by simultaneously handling \textbf{discrete text tokens} and \textbf{continuous image features} under the \textbf{AR} modeling principle. The continuous treatment of visual signals minimizes the information loss while the fully AR formulation renders the characterization of the correlation between modalities straightforward. 
Orthus leverages these advantages through its modality-specific heads---one regular language modeling (LM) head predicts discrete text tokens and one diffusion head generates continuous image features. 
% In practice, Orthus operates on the hidden representation of a pre-trained image variational autoencoder (VAE) for state compression. 
% Unlike fully token-based autoregressive (AR) models, Orthus adopts continuous image representations from a VAE encoder, avoiding information loss caused by Vector Quantization (VQ). 
% Compared with prior studies that unify autoregressive and diffusion in a single transformer, Orthus is pure AR by disentangling diffusion from the backbone, making it more suited to image-text interleaved modeling (because the backbone does not need to accept noisy images during training). 
We devise an efficient strategy for building Orthus---by substituting the Vector Quantization (VQ) operation in the existing unified AR model with a soft alternative, introducing a diffusion head, and tuning the added modules to reconstruct images, we can create an Orthus-base model effortlessly (e.g., within 72 A100 GPU hours). 
Orthus-base can further embrace post-training to craft lengthy interleaved image-text, reflecting the potential for handling intricate real-world tasks. For visual understanding and generation, Orthus achieves a GenEval score of 0.58 and an MME-P score of 1265.8 using 7B parameters, outperforming competing baselines including Show-o and Chameleon. Our code is available at \url{https://github.com/zhijie-group/Orthus}.
\end{abstract}

 % bypasses the need to add noise (whether [mask] tokens or Gaussian noise) to the image before processing it in the transformer, enabling more effective multimodal learning.
 % Specifically, Orthus accepts discrete text tokens and continuous image features as inputs and is composed of a unified transformer backbone and two dedicated heads for text and image decoding respectively. The backbone auto-regressively models the correlation, while the heads yield discrete text tokens and continuous image features, respectively. The heads are trained under typical softmax cross-entropy and diffusion modeling rules. Unlike fully autoregressive (AR) models, Orthus adopts continuous, lossless patch-wise image features from a VAE encoder and enjoys enhanced image-generation quality benefits from Vector Quantization (VQ)-free representations.    
\vspace{-18pt}
\section{Introduction}
\label{sec:intro}

Multimodal models have shown promise in image-to-text and/or text-to-image generation, with LLaVA~\cite{liu2024visual,liu2024llava}, Emu2~\cite{sun2024generative}, and NExT-GPT~\cite{wunext} as popular examples. 
These abilities are essential for handling complex real-world understanding and generation problems. 
Yet, existing approaches can suffer from significant modeling redundancy due to the trivial combination of specialized large models (\eg, CLIP-ViT~\cite{radford2021learning}, Stable Diffusion~\cite{rombach2022high}, and LlaMa~\cite{touvron2023llama,touvron2023llama2}). % are trivially assembled in a system. %, with separate models handling visual understanding and generation. 
% For example, NExT-GPT employs a language model solely for understanding tasks, requiring an external diffusion model for image generation. 
% This can cause modeling redundancy and
Doing so also undermines the benefits brought by cross-modal learning and introduces considerable inefficiency for both training and inference. %, limiting the ability of the learned model to generate mixed-modality contents.

\begin{figure*}[t]
  \centering
   \includegraphics[width=1\linewidth]{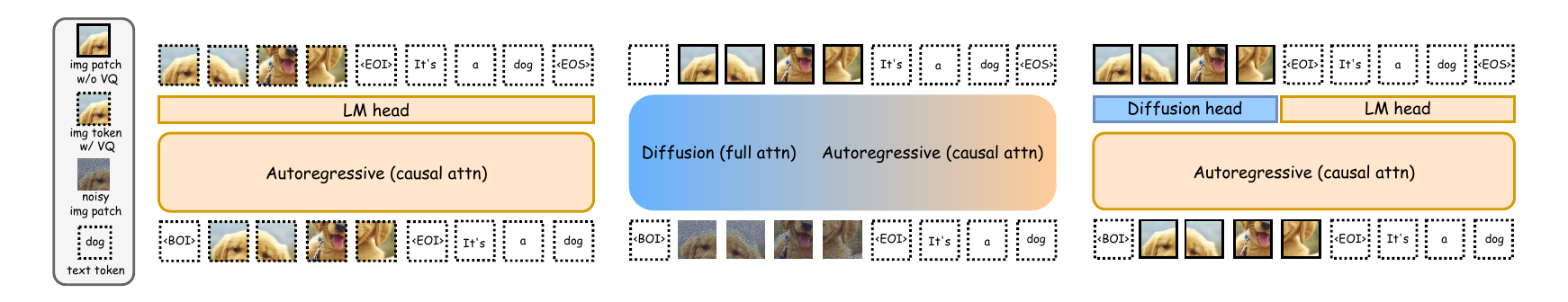}
   \caption{Comparison of existing unified multimodal models with Orthus. 
   \textbf{Left:} \textbf{Fully AR models}~\cite{liu2024worldmodelmillionlengthvideo,team2024chameleon,wang2024emu3} convert visual signals to discrete image tokens via vector quantization for joint modeling with text tokens, but this causes information loss. \textbf{Middle:} \textbf{AR-diffusion mixed models}~\cite{zhou2024transfusion,xie2024show} perform next-token prediction for text generation and image patch denoising for image generation, but the involved noise disturbance on images makes the concurrent image-to-text and text-to-image generation challenging. \textbf{Right:} \textbf{Orthus} operates in a fully AR manner while circumventing vector quantization and noise disturbance to preserve input information and modeling flexibility.}
   \label{fig:intro_work_compare}
\end{figure*}

There is ongoing interest in jointly modeling visual understanding and generation with a unified, compact model. % for concurrent visual understanding and generation. 
One strategy is to map both images and texts to discrete tokens for simple autoregressive (AR) modeling~\cite{liu2024worldmodelmillionlengthvideo,team2024chameleon,wang2024emu3} (left of Figure~\ref{fig:intro_work_compare}).
However, the image tokenizer, often equipped with a vector quantization (VQ) bottleneck, can cause inevitable information loss and easily lead to suboptimal performance on vision tasks concerning high-frequency details (\eg, OCR and human face generation). 
Alternatively, recent works, including Transfusion~\cite{zhou2024transfusion} and Monoformer~\cite{zhao2024monoformer} (middle of Figure~\ref{fig:intro_work_compare}), propose to integrate AR modeling on discrete text tokens and diffusion modeling on continuous image features within a single transformer. 
Nonetheless, the nature of diffusion modeling to process noisy images~\cite{ho2020denoising} makes the joint modeling of image-to-text, text-to-image, and more complicated interleaved image-text challenging. %, eliminating the possibility for mixed-modality interleaved generation.  
% Both of them require feeding noisy images to the transformer backbone during training, inducing a gap between training and inference (the backbone accepts noisy images during training but clean images during inference) and hindering effective cross-modality interdependence modeling. 
% For example, Transfusion~\cite{zhou2024transfusion} identifies a nearly 15\% drop in image captioning (measured by CIDEr) when full-range noise is introduced during training. 

% Furthermore, when training with interleave data (\eg text-image-text-image), this can lead to inconsistency within images and misalignment between image and text.
 % and posing challenges for modeling cross-modality interdependence
% Both Chameleon~\cite{team2024chameleon} and Lumina-Next~\cite{liu2024lumina} have identified that this vector quantization (VQ) bottleneck  
% On the other hand, diffusion models~\cite{esser2024scaling,ho2020denoising,rombach2022high,wu2023tune} in image/video generation capacity have exhibited superior visual generation capabilities than autoregressive. 

This paper proposes Orthus\footnote[1]{Orthus is a loyal two-headed guard dog in Greek mythology.} to bridge the gap. %, establishing a more reliable and efficient way for unified mixed-modality generation. 
Orthus conjoins \emph{lossless continuous image features} and \emph{the unified, cross-modal AR modeling} by decoupling diffusion from the transformer backbone. 
This circumvents the noise disturbance and renders the characterization of the correlation between modalities straightforward, making it more suitable for interleaved image-text modeling.
Specifically, Orthus embeds both discrete text tokens (from an off-the-shelf tokenizer) and continuous patch-wise image features (from a pre-trained variational autoencoder~\cite{vae}) into the same representation space, where an AR transformer is then invoked to model the inter- and intra-modality interdependence. % for effective information abstraction. 
% given a sequence of discrete text tokens and continuous image patches\footnote{Compared with discrete image tokens, we refer to continuous patch-wise features encoded by a VAE as image patches.}, 
% Orthus models the interdependence of text tokens and image patches (both within and across modalities) auto-regressively within a transformer backbone, 
On top of the backbone, Orthus defines two modality-specific heads, with one as the regular \emph{language modeling (LM) head} to predict discrete text tokens and the other as a novel \emph{diffusion head} to craft continuous image features. 
% per-token distribution categorically with a linear head, and per-patch distribution by conditional diffusion with an MLP head. 
% The heads are trained with cross-entropy and diffusion loss respectively on mixed-modality 
% Orthus is exposed to both modalities and optimized with weighted cross-entropy and diffusion loss at each training step and
During inference, Orthus autoregressively predicts the next text token or image patch according to the indication of special transition tokens. % like $<$Begin Of Image$>$ ($<$BOI$>$) and $<$End Of Image$>$ ($<$EOI$>$).

Notably, the investigation into diffusion head for superior image generation draws a striking analogy to the recent masked AR (MAR) approach~\cite{li2024autoregressive}, yet with a focus shift from image-only generation to mixed-modality one. 
% Unlike DiT~\cite{peebles2023scalable} that utilizes diffusion to model the joint distribution of all patches, MAR only uses it to model per-patch distribution while capturing correlation among patches through (masked) autoregression. 
% MAR yields superior image quality due to its VQ-free modeling with diffusion loss. 
On the other hand, our Orthus differentiates from MAR and its variant~\cite{yang2024mmar} in that it characterizes the correlation with fully AR formulation instead of mask-based modeling, which avoids expensive hyperparameter specification and eases the modeling of interleaved data. 
% Drawing insights from MAR's success, Orthus has the potential to outperform 
% AR-based multimodal models in terms of image generation and empirical results also validate this as shown in Table \ref{tab:geneval} and \ref{tab:hps}.

The other important contribution of this work is a super-efficient strategy to build Orthus. % without
% Training a base Orthus is both data-efficient and parameter-efficient, and we find it unnecessary to train it from scratch. 
Inspired by that Orthus differentiates from the representative token-based AR model Chameleon~\cite{team2024chameleon} only in the input embedding modules and output heads, we propose to substitute the VQ operation with a soft alternative and augment the model with an extra diffusion head to instantiate Orthus. 
% Many AR-based unified models, such as Chameleon~\cite{team2024chameleon}, have been trained on millions of image-text pairs, demonstrating strong mixed-modality modeling capacities. Moreover, Orthus and Chameleon share the same nature of AR modeling within the transformer backbone. 
% Accordingly, we propose an efficient training recipe that embraces the pre-trained Chameleon and adapts it to Orthus. 
We tune only the embedding modules and diffusion head (with 0.3B parameters in total) to reconstruct images on a 10k dataset to effortlessly obtain an Orthus base model. 
Orthus-base can further adopt post-training to bolster its ability to model interleaved images and text.
% During adaption, we freeze the transformer backbone to preserve its ability to model correlations between modalities, while only tuning the smoothed vision connector and diffusion head (0.3B parameters in total) with 10k image-only data for distribution matching.

We have performed extensive studies to evaluate Orthus. For mixed-modality understanding and generation, Orthus outperforms the editing-specific model Instruct-pix2pix~\cite{brooks2023instructpix2pix} and exhibits in-context learning capabilities for unseen tasks. Furthermore, Orthus demonstrates a strong ability to generate logically coherent interleaved image-text content with high relevance. For visual understanding and generation, Orthus is substantially superior to Chameleon and Show-o~\cite{xie2024show} across multimodal understanding and generation tasks. %capacity than Chameleon post-trained with the same high-quality data. 
Notably, Orthus achieves a GenEval~\cite{ghosh2024geneval} accuracy of 0.58 and a POPE score of 79.6, even surpassing specialized text-to-image models SDXL~\cite{podellsdxl} and the performant InstructBLIP-13B~\cite{dai2023instructblipgeneralpurposevisionlanguagemodels}. 
% Furthermore, on image understanding tasks like POPE~\cite{li2023evaluating} and GQA~\cite{hudson2019gqa}, Orthus achieves scores of 79.6 and 52.8, outperforming the performant InstructBLIP-13B~\cite{dai2023instructblipgeneralpurposevisionlanguagemodels}. 
% We also showcase the ability of Orthus for mixed-modality generation in both zero-shot scenarios and downstream fine-tuning ones like storybook creation. 

To summarize, our contributions are as follows:
\begin{itemize}
    \item We introduce Orthus for interleaved image-text generation. 
    Orthus models the correlation between modalities through AR principle and generates discrete text tokens and continuous image features with dedicated heads. 
    % Orthus generates discrete text tokens and continuous image features with dedicated heads to effectively model cross-modal correlations through the AR principle.
    % Orthus models the correlation between modalities through the AR principle and generates discrete text tokens and continuous image features with dedicated heads. 
    %, which eliminates information loss while enjoying precise and straightforward modeling of correlation between modalities.
    \item We propose an efficient strategy to build Orthus by exploiting its connection with existing unified AR models, which reduces the cost to merely 72 A100 GPU hours.
    \item Compared to related works such as Chameleon~\cite{team2024chameleon} and Show-o~\cite{xie2024show}, Orthus outperforms them across various visual understanding and generation benchmarks, while also demonstrating extra capabilities in mixed-modality understanding and generation, positioning it as a promising approach for unified multimodal modeling.
\end{itemize}

\section{Related Work}
\label{sec:related}
% \textbf{Visual generation.} ar-based work(+var). diffusion-based work. ar+diffusion based work(MAR, Hart, Fluid). connection with our work.
\noindent \textbf{Visual understanding.}
To enable multimodal large language models (MLLMs) to comprehend modalities beyond text, prior work has introduced methods that leverage pre-trained, modality-specific encoders~\cite{radford2021learning, li2022blip, yu2022coca, chen2024internvl} to generate latent representations for each modality. These representations are then projected into a pre-trained LLM’s input space through trained adapters, allowing for multimodal information alignment within the language model, understanding are handled within the transformer backbone~\cite{liu2024visual, zhu2023minigpt, dai2023instructblipgeneralpurposevisionlanguagemodels,driess2023palm, chen2023pali, liu2024improved, lin2024vila}. This framework allows LLMs to perform complex multimodal tasks while maintaining the language-based reasoning capabilities inherent to their architecture.

\noindent \textbf{Visual generation.}
The generation of visual content has long been a central focus within the deep learning research community~\cite{kingma2013auto, goodfellow2014generative, karras2019style, vahdat2020nvae}. 
% Classical approaches, such as Variational Autoencoders (VAEs)~\cite{kingma2013auto} and Generative Adversarial Networks (GANs)~\cite{goodfellow2014generative}, seek to model complex target distributions directly, aiming to generate samples in a single pass through the neural network. 
Over the past few years, research in visual generation has focused on decomposing visual signals in a more sophisticated manner and generating them iteratively. Diffusion models~\cite{sohl2015deep,ho2020denoising,song2020score,dhariwal2021diffusion,rombach2022high,peebles2023scalable,esser2024scaling} transform generation into a reverse diffusion process from noise to data, gradually refining an initial noise input through a series of denoising steps. While another line of work aims to emulate the success of AR modeling from language modeling within the visual domain~\cite{parmar2018image, razavi2019generating, ramesh2021zero, yu2023language, sun2024autoregressive}. Specifically, images are first transformed into a sequence of vector-quantized tokens~\cite{van2017neural, esser2021taming, tian2024visual, yu2024image}, after which AR modeling is then performed on the discrete-valued token space~\cite{touvron2023llama}. To mitigate generation quality degradation caused by information loss during the VQ process, MAR replaces the per-token categorical distribution modeling with a diffusion procedure~\cite{li2024autoregressive, fan2024fluid}. Our proposed method generalizes MAR to cross-modality generation.

% HART~\cite{tang2024hart} utilizes a hybrid tokenizer and simultaneously models the discrete per-token distribution and the continuous residual components with a diffusion model.

\noindent \textbf{Unified visual understanding and generation.} 
To enable a model to possess both understanding and generation capabilities, one kind of approach aims to connect LLMs with multimodal adapters and diffusion decoders~\cite{sun2023emu, ge2024seed, ye2024x}. However, using multiple distinct components can lead to redundancy and inefficient information use. Consequently, recent studies seek to leverage a single transformer for unified understanding and generation. 
A straightforward approach is to apply vector quantization to continuous visual signals to enable visual tokens, like discrete text tokens, to be trained within a unified token space using cross-entropy loss. Representative works are LWM~\cite{liu2024world}, Chameleon~\cite{team2024chameleon}, Anole~\cite{chern2024anole}, and VILA-U~\cite{wu2024vila}. Alternatively, some works have explored combining AR with diffusion modeling.
Show-o~\cite{xie2024show} unifies AR and discrete diffusion modeling for multimodal understanding and generation within one single transformer. 
Transfusion~\cite{zhou2024transfusion} and Monoformer~\cite{zhao2024monoformer} train one shared transformer for both discrete text autoregression and continuous image diffusion. Our proposed method circumvents the potential information loss caused by quantization and noise disturbance.

% Chameleon~\cite{team2024chameleon} and Anole~\cite{chern2024anole} use a unified token space to reason over interleaved image and text sequences. VILA-U~\cite{wu2024vila} employs contrastive learning to align the visual encoder with its textual inputs, then a unified AR next-token prediction framework is used to conduct all understanding and generation tasks. These methods all apply vector quantization to continuous visual signals to enable visual tokens, like discrete text tokens, to be trained using cross-entropy loss. 
% LWM~\cite{liu2024world} extends its transformer's context window to 1 million tokens to process both video and text sequences.
% The most related is a concurrent work named MMAR~\cite{yang2024mmar}, which adopts a masked random order for image patch generation, while our approach maintains strict AR causal modeling.

% ar-based(LWM, Chameleon, Anole, Vila-U), ar-diffusion(show-o, transfusion, monoformer), connection with our work.
% MM-Interleaved, SEED-X, EMU, DreamLLM
\section{Preliminary}
Unified multimodal modeling aims to cope with a blend of images and texts with a single compact model% and generate a flexible blend of images and text
~\cite{liu2024worldmodelmillionlengthvideo,team2024chameleon,wang2024emu3,zhou2024transfusion,xie2024show}. 
The model usually includes a vision autoencoder, specified with an encoder $E$ and a decoder $D$, a text tokenizer, and a transformer network~\cite{vaswani2017attention}. 
The encoder $E$ is used to map the input image to a sequence of patch-wise features $V := [\vv_1, \ldots, \vv_n ]$, $\vv_i \in \mathbb{R}^{d_v}$ for effective information compression, where $d_v$ is the feature dimension and $n$ is the number of patches. 
% To map images and texts to the same latent representation space for joint modeling, the model utilizes $E$ to map images to a sequence of patch-wise features $\bm{Z} := \{ \vz^i \in \mathbb{R}^{D} \}_{i=1}^n$ ($D$ denotes the dimension of the latent space and $n$ denotes the sequence length depending on the image size); 
The text tokenizer maps the input text into a sequence of text tokens $U := [u_1, \ldots, u_m ]$ with $m$ as the sequence length. 
The transformer is then asked to process $U$ and $V$ simultaneously to yield meaningful outputs, which can be then detokenized as texts or decoded by $D$ to produce images. 
There are primarily two strategies for the learning of the transformer, detailed as follows. 
% The following outlines two primary approaches to unified multimodal models: fully AR models and AR-diffusion mixed models.

\noindent\textbf{Fully AR models.} 
Observing that the AR principle excels in the generative modeling of discrete content, seminal works,
% To enable fully token-based AR modeling, representative works like 
including LWM~\cite{liu2024world} and Chameleon~\cite{team2024chameleon}, propose to leverage the Vector Quantization (VQ)~\cite{van2017neural} technique to transform the continuous image features $V$ as discrete tokens to enable a fully AR modeling of the mixture of images and texts. 
% as the vision autoencoder to enable image tokens from $\bm{Z}$. 
Specifically, VQ introduces a set of $K$ codes $\{\vc_j \in \mathbb{R}^{d_v}\}_{j=1}^K$ and solves the following problem for continuous-to-discrete transformation:
\begin{equation}
\label{eq:argmax}
   \tilde{v}_i = \underset{j \in \{1, \ldots, K\}}{\mathrm{arg\,min}} \, d(\vv_i, \vc_j)\;\, \text{for}\,\,i=1, \ldots, n,
\end{equation}
where $d(\cdot, \cdot)$ is a distance metric. 
% Conversely, the discrete tokens $\{x_i\}_{i=1}^{n}$ can be detokenized back to reconstruct the origin image by passing the associated latent codes $\{C_{x_i}\}_{i=1}^{n}$ through the decoder. Texts are tokenized into discrete tokens through the text tokenizer using subword segmentation algorithms, such as Byte-Pair Encoding (BPE)~\cite{sennrich2015neural}.
% A fully AR model comprises an image tokenizer, a text tokenizer, a transformer backbone, and a language modeling (LM) head. The image tokenizer is often instantiated as a Vector Quantized-variational autoencoder (VQ-VAE)~\cite{van2017neural} with an encoder, a decoder, and a codebook $C \in \mathbb{R}^{K\times d_c}$, where $K$ is the size of the codebook and $d_c$ is the dimensionality of each latent code.
% To process interleaved images and texts, they are first tokenized into a sequence of discrete tokens $\{x_i\}$ (each $x_i$ is an integer). Specifically, images are first passed through the encoder of VQ-VAE producing continuous patch-wise features. These patches are then arranged in raster order, forming a sequence of image patches $\{\vz_i\}_{i=1}^n$ ($n$ denotes the sequence length depending on the image size), and the discrete tokens are then calculated by the nearest neighbor look-up using the codebook $C$ as shown in Equation \ref{eq:argmax}:

Let $\tilde{V} := [\tilde{v}_1, \ldots, \tilde{v}_n]$ denote the discrete image tokens. 
The fully AR model embeds both $\tilde{V}$ and $U$ as $d_e$-dim features. % with separate embedding weights.
Specifically, the embedding corresponding to $\tilde{v}_i$ is
\begin{equation}
\label{eq:embed}
    \vh_i = \sum_j \vw_j \mathbbm{1}_{\tilde{v}_i = j}, 
\end{equation}
where $\{\vw_j \in \mathbb{R}^{d_e}\}_{j=1}^K$ refer to the embedding weights. 
The embeddings for text tokens can be similarly gained, yet with another set of embedding weights. 
% Given the token sequence $\{u_1, \ldots, u_n, t_1, \ldots, t_m\}$, this type of model embeds them into vectors and processes the embedded vectors by $f_\theta$ via 
The transformer then processes these embeddings with \emph{causal attention}, where the output head naturally yields the prediction of the next token. 
For training, the objective is simply the AR loss. 
% Following common practice, the training is performed under the teacher-forcing AR loss. 

% $L_\theta$ to generate subsequent tokens categorically.

% $p(x^1,\ldots,x^l)=$ \resizebox{.24\hsize}{!}{$\prod_{i=1}^l p(x^i|\vx_{<i})$}, 
% where $\vx_{<i} = \{x^1, \ldots, x^{i-1}\}$. To parameterize $p(\cdot|\vx_{<i})$, the model explicitly represents it as a categorical distribution calculated by the softmax activation on a set of logits predicted by $f_\theta$ followed by the language modeling (LM) head $L_\theta$:
% \begin{equation}
% \label{eq:ar_para}
%     p_{\theta}(\cdot|\vx_{<i}) = \text{softmax}(L_\theta(f_\theta(\vx_{<i}, \mM))),
% \end{equation}
% where $\mM$ is the casual mask matrix.
% The model can then be trained with AR loss in Equation \ref{eq:chameleon_train} and generates subsequent tokens by sampling from $p(\cdot|\vx_{<i})$.
% \begin{equation}
% \label{eq:chameleon_train}
%     \mathcal{L}_{\text{ar}} = -\underset{i=1}{\sum}\log p_{\theta}(x^i|\vx_{<i}).
% \end{equation}

Despite being simple, the fully AR models can suffer from information loss~\cite{liu2024lumina,team2024chameleon}, because VQ makes the transformer unable to directly look at the image features $\vv_i$. 

\begin{figure*}[t]
  \centering
    \includegraphics[width=0.88\linewidth]{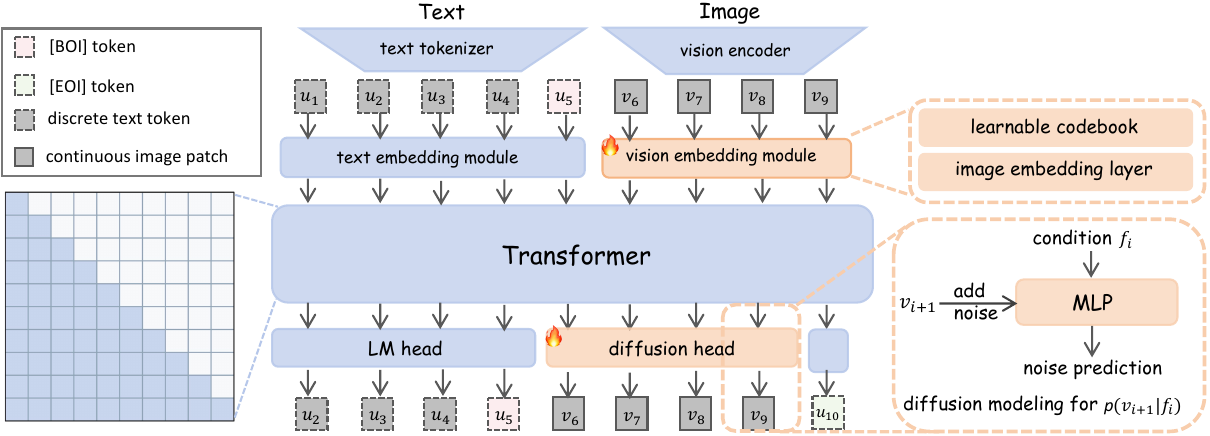}
   \caption{Architecture of Orthus. Orthus is composed of a text tokenizer, a vision autoencoder, two modality-specific embedding modules, a transformer backbone, and two modality-specific heads. 
   % Given images and texts, 
   Orthus tokenizes texts into discrete text tokens and encodes images into continuous patch-wise features. 
   They are then embedded as a sequence of vectors and processed by the transformer backbone with causal attention, generating a sequence of output vectors. 
   The vectors are routed to modality-specific heads, with the \emph{LM head} to predict the next text token categorically and the \emph{diffusion head} to predict the next image patch feature through conditional diffusion modeling.}
   \label{fig:model_arc}
\end{figure*}

\noindent\textbf{AR-diffusion mixed models.} Another line of unified multimodal models is AR-diffusion mixed models~\cite{zhou2024transfusion,xie2024show,zhao2024monoformer}, which integrates diffusion modeling on images~\cite{ho2020denoising,peebles2023scalable} and AR modeling on text within a shared transformer. 
Take Transfusion~\cite{zhou2024transfusion} for example, 
% it adopts diffusion modeling for $V$ and AR modeling for $T$. During training, it first 
its inputs are a noisy version of 
% samples noise from $\mathcal{N}(\vzero, \mathbf{I})$ in the form of $n$ patches and 
the image features $V$, denoted as $\bar{V}$, 
% according to a noise schedule~\cite{ho2020denoising} to get the noise-corrupted $V_t$. $V_t$ 
and the text tokens $U$. 
To facilitate the simultaneous processing of $\bar{V}$ and $U$, the attention mask of the transformer adopts a unique configuration---with a full-attention structure among $\bar{V}$ and a causal structure among $U$. 
% via an expanded causal attention mechanism, where patches of $V_t$ from the same image are conditioned on each other via \emph{full attention}. 
Then, the outputs from $\bar{V}$ are directed to an output projector to predict the noise on $\bar{V}$, whereas the outcomes linked to $U$ are channeled to an LM head for next-token prediction. 
The training objective is the combination of AR loss and denoising loss with a balancing factor. 
During inference, the model operates as an AR model to generate texts and as a diffusion model to craft images, with special tokens indicating mode switching.

However, diffusion modeling inherently requires feeding noisy inputs to the model, hindering joint modeling of visual understanding (requiring clean images) and generation (requiring noisy ones). 
For example, Transfusion identifies a nearly 15\% performance drop in image captioning when full-range noise is introduced during training.
% Also clarify the issues of such models. 
% specify the details for such models, including input, output, and learning objective.

\section{Method}
\label{sec:method}
We introduce Orthus to address the issues of existing works. This section begins with an overview of Orthus and then elaborates on an efficient training recipe for Orthus. 
We will also illustrate a post-training pipeline of Orthus. %to fully unlock the multimodal understanding and generation capacity of Orthus. 

\subsection{Overview of Orthus}
% Pre-trained on 1.4B text-image pairs and interleaved text-image data, Chameleon demonstrates strong multimodal capabilities. However, its performance in visual understanding and generation is limited by the quantization bottleneck. Alternatively, we propose Orthus, a new unified framework for effective multimodal learning that breaks this bottleneck and employs diffusion modeling to enhance visual generation quality.

As shown in Figure~\ref{fig:model_arc}, Orthus directly takes the continuous image features $V$ and discrete text tokens $U$ as input, which avoids the pathologies caused by the quantized image features $\tilde{V}$ or noisy image features $\bar{V}$. 
$U$ and $V$ are embedded into the $d_e$-dim representation space with a differentiable vision embedding module (detailed in the next subsection) and the aforementioned discrete embedding module respectively. 
% utilizes the same process as Equation \ref{eq:embed} to embed $T$ and embeds $V$ via a soft vision embedding module, which is illustrated explicitly in Section \ref{sec:soft_embed}. 
Subsequently, the embeddings are fed into the transformer backbone with \emph{causal attention} for the modeling of both inter- and intra-modality interdependence. 
Given the output states of such a backbone contain enough information about the multimodal context, Orthus sends them to two modality-specific heads---a diffusion head and an LM head---to predict the next image patch or the next token. 

Specifically, let $\vf_i$ denote the output state corresponding to the input image feature $\vv_i$ and $\epsilon_\theta$ denote the diffusion head employed by Orthus with parameter $\theta$. 
The goal of $\epsilon_\theta$ is to predict for the next patch feature $\vv_{i+1}$ conditioning on $\vf_i$. 
% generating output vectors $[\vc_1, \ldots, \vc_l]$, $\vc_i \in \mathbb{R}^{d_e}$ and $l$ is the sequence length. 
% Each $\vc_i$ extracts the information of preceding tokens $t_{<i}$ and features $\vz_{<i}$ and is routed to the modality-specific head---a \emph{LM head} $H^l_\theta$ or a \emph{diffusion head} $H^d_\theta$---to predict the distribution of next text-token $t_i$ or image-feature $\vv_i$, respectively. 
% To be concrete, $H^l_\theta(\vc_i)$ predicts a categorical distribution to sample $t_i$ as AR modeling, while $H^d_\theta$ parameterizes a probability distribution of $\vv_i$ conditioned on $\vc_i$, that is, $p(\vv_i|\vc_i)$ in a diffusion way. 
According to common practice~\cite{ho2020denoising,dhariwal2021diffusion}, the learning objective for the diffusion head can be formalized as:
% The standard diffusion loss~\cite{ho2020denoising,dhariwal2021diffusion}:
\begin{equation}
\label{eq:diff_loss}
    \mathcal{L}_{\text{diff}} = \E_{\bm{\epsilon},t}[\Vert \bm{\epsilon}-\epsilon_\theta(\sqrt{\overline{\alpha}_t} \vv_{i+1} + \sqrt{1-\overline{\alpha}_t}\bm{\epsilon}, t,\vf_i)\Vert_2^2],
\end{equation}
where $\bm{\epsilon}\sim \mathcal{N} (\vzero, \mathbf{I})$ is a Gaussian noise and $t$ is a randomly sampled timestep.
$\overline{\alpha}_t$ follows a pre-defined noise schedule~\cite{ho2020denoising}. 
In practice, $\epsilon_\theta$ can be a shallow multilayer perception (MLP) with three inputs (the condition $\vf_i$, the scalar timestep $t$, and the noisy state). 
On the other hand, the LM head remains the compact linear projection followed by a softmax transformation to yield the predictive probability of the next token over the entire vocabulary.

\subsection{An Efficient Strategy for Constructing Orthus-base} 
The differences between Orthus and fully AR models exist in the vision embedding module and the output head.
Given that pre-training a multimodal model from scratch can be frustratingly costly but the fully AR models like LWM~\cite{liu2024world} and Chameleon~\cite{team2024chameleon} are readily accessible from the open-source community, we are naturally interested in deriving Orthus based on them at a minimal expense.
This section elaborates on a hard-to-soft adaptation trick and an efficient training strategy to enable this. 

% we propose an efficient training strategy to build Orthus by adapting full AR models.
\label{sec:soft_embed}
% Given the same nature of autoregressive modeling of interleave sequence with Chameleon, we develop a recipe to adapt it to Orthus. This process fully embraces pre-trained Chameleon's capacity and avoids the expensive computation overhead of large-scale pre-training. 

\noindent\textbf{Differentiable vision embedding module.} 
It is easy to note that the embedding yielded by Equations~\ref{eq:argmax} and~\ref{eq:embed} can be equivalently obtained via a softmax-based transformation
\begin{equation}
\label{eq:softmax}
\vh_i = \sum_j \vw_j \frac{e^{-d(\vv_i, \vc_j)/\tau}}{\sum_{k=1}^K e^{-d(\vv_i, \vc_k)/\tau}},
\end{equation}
with $\tau \to 0$. 
Increasing $\tau$ gradually from $0$ then naturally lifts the information bottleneck from the image features $\vv_i$ to the model outputs $\vf_i$, while rendering the reuse of the pre-trained weights and codes of fully AR models possible. 
% Combining Equations~\ref{eq:argmax} and~\ref{eq:embed} yields the following embedding process of image features $V$ in existing fully AR models:
% \begin{equation}
%     \vh_i = \sum_j \vw_j \mathbbm{1}_{\underset{k \in \{1, \ldots, K\}}{\mathrm{arg\,min}} \, d(\vv_i, \vc_k) = j} 
% \end{equation}
% Given that the embedding process (Equation \ref{eq:argmax} and \ref{eq:embed}) in fully AR models compresses the spatial of $V$ and causes information loss due to the argmin operation in the VQ step, we develop a method to break this bottleneck. S
% pecifically, we substitute VQ into a soft alternative and map $\vv_i$ to embedded vectors $\vh_i$ through a smoother process:
% where $\tau$ is the temperature and when it is set to $0$, Equation \ref{eq:softmax} degrades to the embedding process in fully AR models. 
% This replacement enhances the information flow from $\vv_i$ to the predicted next tokens and features. 
This way, the codes $\{\vc_j\}_{j=1}^K$ also become a part of the input module, so we can leverage gradients to directly push them to adapt to the multimodal learning tasks. 
This contradicts fully AR models which froze the codes during training. 
% it facilitates gradient propagation back to the codebook $\{\vc_j\}_{j=1}^K$ through the loss function, allowing 
% $\{\vc_j\}_{j=1}^K$ to be trained for visual understanding and generation, rather than being frozen and solely optimized for visual reconstruction. Consequently, the trainable $\{\vc_j\}_{j=1}^K$ combined with $\{\vw_j\}_{j=1}^K$ can serve as a soft vision embedding module.
% Rethinking the process in Chameleon that maps the latent representations $z$ to a sequence of vectors $\{h_i\}_{i=1}^{1024}$ aligned with the transformer's space, we can conceptualize it as a discrete projection process with quantization. In this framework, the codebook $\mathcal{C}$ combined with the embedding layer corresponding to image tokens $\mathcal{W}_{\text{image}}$ functions as a discrete connector. Specifically, the continuous latent image patch $z_i$ is projected through the following process:
% \begin{equation}
%     x_i = \underset{j \in K}{\mathrm{argmax}}(-\| z, \mathcal{C}_j \|_2); h_i = \mathcal{W}_{\text{image}}e_i,
% \end{equation}
% where $e_i$ denotes the one-hot encoding of $x_i$. 
% We can smooth this process by 

% \noindent\textbf{Modality-specific heads.} For token-wise distribution modeling, we directly leverage the linear LM head corresponding to text tokens in Chameleon. For patch-wise distribution modeling, we substitute the origin linear LM head corresponding to image tokens to a lightweight (49M parameters) diffusion MLP. 
% , generating a sequence of output vectors $\{c_i\}$. Each output $c_i$ captures information of preceding tokens $x_{<i}$ and

\noindent\textbf{Training strategy.} 
With the above trick, we start with a pre-trained fully AR model, transform its input module into a differentiable one, and introduce an output diffusion head to initialize Orthus. 
These modifications primarily focus on the visual part, thus we recommend fine-tuning the initialized model on a collection of images. 
In particular, we input only the image into Orthus to acquire the hidden states $\vf_i$ and utilize the diffusion loss in Equation~\ref{eq:diff_loss} to recover the next patch to train the vision embedding module and diffusion head. 
The temperature $\tau$ is set to 1 during training.

Initialized with the typical Chameleon-7B~\cite{team2024chameleon}, Orthus can acquire image processing capabilities while preserving the text generation capacity after 9-hour training on 10k high-quality images~\cite{laion-coco} using 8 A100 GPUs. We designate this model as Orthus-base, a pre-trained model capable of generating continuous image features and discrete text tokens.
% which serves as a pre-trained model for further 
% Figure~\ref{fig:interleave} shows its image and text (as well as mixed-modality) generation prowess.

Although the decoder in the VQ-VAE~\cite{van2017neural} used by Chameleon can reconstruct the raw image pixels given the patch-wise features $V$ to some extent, it can be suboptimal due to the quantization-aware training. 
To address this, we advocate further tuning its decoder to reconstruct high-quality images directly based on $V$. 
The comparison between the capacity of the original VQ-VAE and ours is exhibited in Appendix \ref{apped:aoe}.

% In practice, we adapt the representative full AR model Chameleon-7B~\cite{team2024chameleon} to construct Orthus. During adaptation, we replace $\{\vw_j\}_{j=1}^K$ in Chameleon to the soft vision embedding module and augment a diffusion MLP head to instantiate Orthus. 

% Since we only need to fit the conditional diffusion probability per image patch for the adaption of Orthus, 
% we select 10k high-quality image-only data~\cite{laion-coco} and 
% only tune the soft vision embedding modules and the augmented diffusion MLP with diffusion loss in Equation \ref{eq:diff_loss} to reconstruct images. 

% After training for around 9 hours on 8 A100 GPUs, we build a base Orthus which inherits multimodal understanding and generation capacity of pretrained Chameleon.
% To construct a continuous autoencoder compatible with the pre-trained Chameleon, we freeze the encoder of Chameleon's VQ-VAE, drop the quantization step, and finetune the decoder only to transform it into a continuous autoencoder as in ~\cite{tang2024hart}.

\subsection{Multimodal Post-training}
Orthus-base can be further post-trained to unlock its potential for interleaved image-text modeling in complex downstream tasks. These include generating text (\eg, visual question answering), images (\eg, image editing), or even both images and text (\eg, storybook generation) from mixed-modality inputs. For example, given $[V, U]$ as user input and $[V, U, V, U]$ as model output, we surround image features $V$ with the embeddings of special tokens \texttt{[BOI]} and \texttt{[EOI]} before the concatenation with $U$. A \texttt{[SEP]} token is used to separate user input and model output in each conversation. 
% Following the typical training procedure of multimodal models~\cite{xie2024show,wang2024emu3nexttokenpredictionneed}, we further refine Orthus-base by fine-tuning it on high-quality text-image pairs and instructional data. 
% Following the input format of Chameleon, we surround image features $V$ with the embeddings of special tokens \texttt{[BOI]} and \texttt{[EOI]} before the concatenation with text embeddings. 
% A \texttt{[SEP]} token is used to separate user input and model output in each conversation. 

Let $\mathcal{L}_{\text{ar}}$ denote the AR loss on the text tokens.
% objective $\mathcal{L}_{\text{ar}}$ to per text-token and the diffusion objective $\mathcal{L}_{\text{diff}}$ in Equation \ref{eq:diff_loss} to per image-patch. 
% Consequently, t
The entire training objective of Orthus is then $\mathcal{L}_{Orthus} = \mathcal{L}_{\text{ar}} + \lambda \mathcal{L}_{\text{diff}}$, where $\lambda$ is a balancing coefficient. All parameters except for those of the vision autoencoder are tuned. 
Hereinafter, we will denote the model trained following this objective as Orthus, distinguishing it from Orthus-base. 
% This unified training loss can also be used to train an Orthus from scratch, which we leave as our future work.
% \footnote{Due to the autoregressive nature, $\mathcal{L}_{\text{diff}}$ is applied when the input is \texttt{[BOI]} token to predict the first image patch and $\mathcal{L}_{\text{ar}}$ is applied when the input is the last image patch to predict \texttt{[EOI]}.}

% \subsection{Inference}
During inference, Orthus alternates between \emph{next-token predition} and \emph{next-patch prediction} to seamlessly generate interleaved texts and images. 
% During \emph{next-token prediction}, the output vector $\vf_i$ is routed to $\epsilon_\theta$ to sample discrete tokens. 
When \texttt{[BOI]} is sampled during the next-token prediction process, the algorithm moves to \emph{next-patch prediction}. %
% and the output vector $\vf_i$ is routed to $\epsilon_\theta$ as a condition input, while a pure noise patch $\vv_i^T$ is fed into it simultaneously as input, predicting the noise accumulated at timestep $T$ and denoising $\vv_i^T$ to $\vv_i^{T-1}$ by removing the corresponding noise proportion~\cite{ho2020denoising,dhariwal2021diffusion}. 
% After denoising $T$ steps, the clean patch is appended to the input sequence. 
Once a fixed number of $n$ image patches are generated, \texttt{[EOI]} is appended and the algorithm switches back to \emph{next-token prediction}.

% then scaled according to the noise schedule, 
\section{Experiments}
\label{sec:exp}
In this section, we evaluate Orthus's performance in interleaved image-text modeling as well as visual understanding and generation. Both quantitative and qualitative results demonstrate the effectiveness of Orthus. 
\subsection{Implementation Details}
We implement the diffusion head as an MLP consisting of 3 residual blocks, each sequentially applying AdaLN~\cite{peebles2023scalable}, a linear layer (width of 1536 channels), SiLU activation, and another linear layer. The condition vector $\vf_i$ is added to the diffusion time embedding, which is then incorporated through AdaLN. The diffusion noise schedule is linear following~\cite{rombach2022high}, with 1000 steps at training time. $\lambda$ is set to $100$ to balance the order of magnitude between $\mathcal{L_\text{diff}}$ and $\mathcal{L_\text{ar}}$ during post-training. During inference, we use greedy decoding to generate text. For image generation, we adopt the DDIM~\cite{songdenoising} sampler with 100 steps. We employ classifier-free guidance (CFG)~\cite{ho2022classifier} with the scale set to 5 during sampling. All images are generated at a resolution of 512$\times$512. More training and evaluation details are provided in Appendix~\ref{apped:train_detail}.
\\

\subsection{Interleaved Image-Text Generation}
\begin{figure*}[t]
\centering
\includegraphics[width=0.93\textwidth]{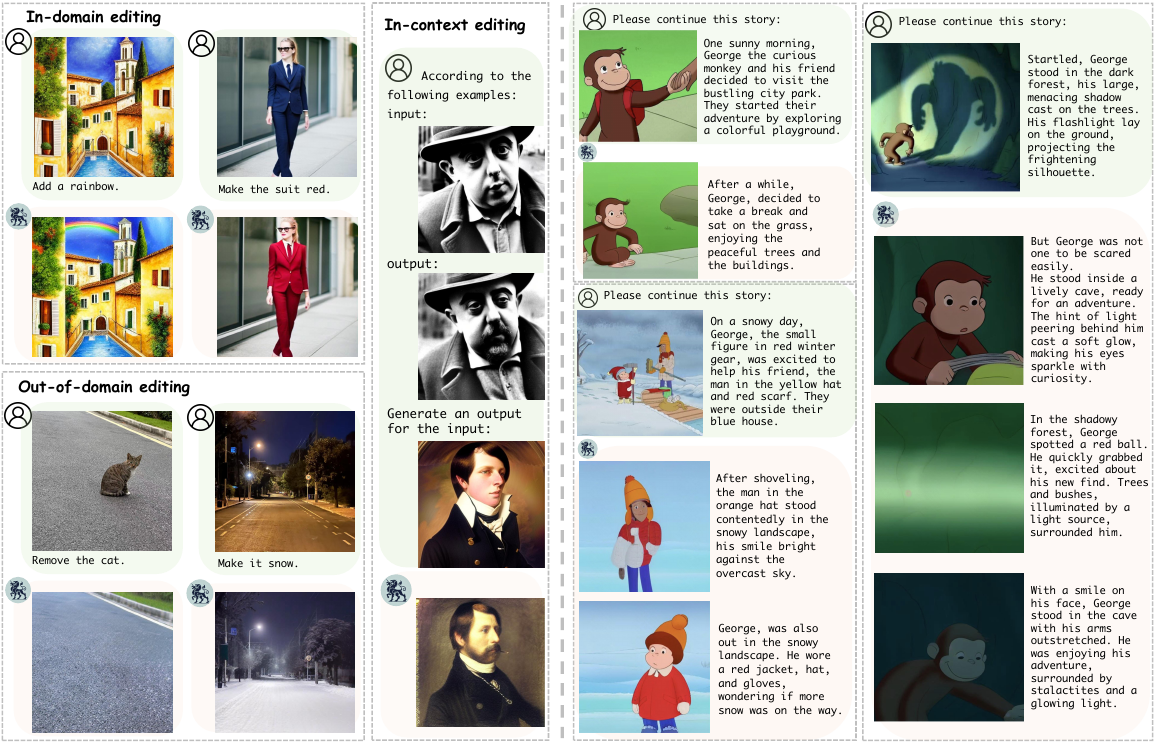} 
\vspace{-6pt}
\caption{Qualitative results on mixed image-text understanding and generation of Orthus. \textbf{Left:} Image editing results after fine-tuned on Instruct-Pix2Pix~\cite{brooks2023instructpix2pix}. Notably, Orthus exhibits \emph{in-context learning} capacity by performing image editing successfully when provided with examples rather than explicit instructions, which is not included in the training dataset.  \textbf{Right:} Interleaved storybook creation results after finetuned on the StoryStream~\cite{yang2024seed} dataset. Results show that Orthus excels in generating logically coherent interleaved image-text with high relevance.}
\label{fig:interleave}
\vspace{-10pt}
\end{figure*}
Compared to existing unified models, such as Janus-series~\cite{wu2024janus, ma2024janusflow}, that focus exclusively on visual understanding and generation, we investigate Orthus's flexibility and extensibility to model interleaved images and text on two representative downstream tasks: image editing and storybook generation.
% In this section, we investigate the mixed-modality generation capabilities of Orthus. We first show that Orthus-base can perform mixed-modality generation, as demonstrated in the left of Figure \ref{fig:interleave}.
% Orthus-base generates a coherent sequence of interleaved image-text segments based on instructions from users in a zero-shot manner. 

\begin{table}[t]
    \centering
    \vspace{-6pt}
    \caption{Comparisons of CLIP similarities~\cite{ruiz2023dreambooth, gal2022stylegan} between editing-specific diffusion models and Orthus on the test dataset of Instruct-Pix2Pix.}
    \vspace{6pt}
    \setlength{\tabcolsep}{6pt}
    \begin{tabular}{lccc}
    \toprule
     \textbf{Model} & \textbf{-T$\uparrow$} & \textbf{-I$\uparrow$} & \textbf{-D$\uparrow$}  \\
    \midrule
    PnP~\cite{tumanyan2023plug} & 0.156 & 0.76 & 0.023 \\
    SDEdit~\cite{mengsdedit} & 0.229 & 0.84 & 0.047 \\
    I-Pix2Pix~\cite{brooks2023instructpix2pix} & 0.233 & \textbf{0.88} & 0.045 \\
    \textbf{Orthus (Ours)} & \textbf{0.238} & 0.87 & \textbf{0.049} \\ 
    \bottomrule
    \end{tabular}
    \label{tab:edit}
    \vspace{-13pt}
\end{table}
\textbf{Image-Text $\boldsymbol{\rightarrow}$ Image.}  We compare the performance of Orthus with an editing-specific diffusion model after training Orthus-base on the 400k Instruct-pix2pix~\cite{brooks2023instructpix2pix} training dataset. Table~\ref{tab:edit} shows that the images edited by Orthus align well with both the given instruction and the input image, performing comparably to or even surpassing the editing-specific diffusion model. Moreover, Figure~\ref{fig:interleave} illustrates Orthus's strong generalization ability to edit images in zero-shot real-image domains. Notably, Orthus exhibits \emph{in-context learning capacity} as a unified multimodal model: when provided with examples instead of explicit instructions that do not match the formats seen during training, it successfully completes the task. This highlights Orthus's strong capability for interleaved data modeling and its great potential as a foundation multimodal model. More editing examples and comparisons are provided in Appendix~\ref{apped:edit}.

\textbf{Image-Text $\boldsymbol{\rightarrow}$ Image-Text-Image-Text.} To further validate Orthus's superiority in modeling interleaved data, we fine-tune Orthus-base with the unified learning objective on the StoryStream~\cite{yang2024seed} dataset, which includes a collection of images and corresponding narratives from cartoon series. 
As shown in Figure \ref{fig:interleave}, after training, Orthus can generate contextually consistent scenes paired with narrative text given an initial image-text pair and the instruction ``Please continue this story.''
% Orthus generates contextually consistent scenes paired with narrative text. 
Notably, there is a strong alignment between images and text (e.g., the smile on the monkey's face) as well as consistent detail across images (e.g., the boy's orange hat and red scarf). These results highlight Orthus’s ability to generate long sequences of contextually relevant images and text. This capability opens up potential applications such as report generation, educational content creation, and other tasks requiring seamless mixed-modality content generation.
\begin{table*}[t]
    \caption{\textbf{Evaluation on visual understanding benchmarks.} Und. and Gen. denote ``understanding'' and ``generation'', respectively. Models using external pre-trained diffusion models are marked with * and $\text{Chameleon}^\dagger$ is post-trained with the same dataset as Orthus. The results in \textbf{bold} and \underline{underline} are the best and second-best results, respectively. The results correspond to the exact match accuracy.}
    \vspace{6pt}
    \centering
    \setlength{\tabcolsep}{3.3pt}
    \begin{tabular}{llccccccc}
    \toprule
    \textbf{Type} & \textbf{Model} & \textbf{\# Params}  & \textbf{POPE}$\uparrow$ & \textbf{MME-P}$\uparrow$ & \textbf{VQAv2}$\uparrow$ & \textbf{GQA}$\uparrow$ & \textbf{MMMU}$\uparrow$ \\
    \midrule
    \multirow{6}{*}{Und. Only}& LlaVa~\cite{liu2024visual} & 7B & 76.3 & 809.6 & - & - & - & \\
    & LlaVA-v1.5~\cite{liu2024improved} & 7B & 85.9 & 1510.7 & 78.5 & 62.0 & 35.4 & \\ 
    & InstructBLIP~\cite{dai2023instructblipgeneralpurposevisionlanguagemodels} & 7B & - & - & - & 49.2 & - & \\
    & Qwen-VL-Chat~\cite{bai2023qwenvlversatilevisionlanguagemodel} & 7B & - & 1487.5 & 78.2 & 57.5 & - & \\
    % & mPLUG-Owl2~\cite{ye2024mplug} & 7B & 85.8 & 1450.2 & 79.4 & 56.1 & - & \\
    & Emu3-Chat~\cite{wang2024emu3} & 8B & 85.2 & 1243.8 & 75.1 & 60.3 & 31.6 & \\
    % & IDEFICS-9B~\cite{idefics} & 9B & - & - & - & 50.9 & - & \\
    & InstructBLIP~\cite{dai2023instructblipgeneralpurposevisionlanguagemodels} & 13B & 78.9 & 1212.8 & - & 49.5 & - & \\
    \midrule
    \multirow{8}{*}{Und. and Gen.} & $\text{Emu}^*$~\cite{sun2023emu} & 13B & - & - & 52.0 & - & - & \\
     & $\text{NExT-GPT}^*$~\cite{wunext} & 13B & - & - & \textbf{66.7} & - & - & \\
    % & $\text{SEED-X}^*$~\cite{ge2024seed} & 17B & 84.2 & 1435.7 & - & 47.9 & 35.6 & \\
    & Gemini-Nano-1~\cite{team2023gemini} & 1.8B & - & - & 62.7 & - & 26.3 & \\
    & Show-o~\cite{xie2024show} & 1.3B & 73.8 & 948.4 & 59.3 & 48.7 & 25.1 & \\
    & LWM~\cite{liu2024world} & 7B & 75.2 & - & 55.8 & 44.8 & - & \\
    & $\text{Chameleon}^{\dagger}$ & 7B & 77.8 & 1056.9 & 57.8 & 49.6 & 26.7 & \\
    & \textbf{Orthus (Ours)} & 7B & \textbf{79.6} & \textbf{1265.8} & \underline{63.2} & \textbf{52.8} & \textbf{28.2} & \\
    \bottomrule
    \end{tabular}
    \label{tab:mmu}
\end{table*}
\subsection{Visual Understanding and Generation}
\label{sec:mmu}
In this section, we validate the effectiveness of Orthus on visual understanding and generation by post-training Orthus-base with a mixture of LlaVA-v1.5-665K~\cite{liu2024visual} and high-quality text-to-image data (JourneyDB~\cite{sun2024journeydb} and LAION-COCO-aesthetic~\cite{laion-coco} recaptioned from ShareGPT-4v~\cite{chen2023sharegpt4v}). We also fine-tune pre-trained Chameleon~\cite{chern2024anole} with the same mixed dataset as Orthus to provide an apple-to-apple baseline.
% \textbf{Benchmarks.} For visual understanding, we evaluate Orthus on POPE~\cite{li2023evaluating}, MME~\cite{fu2024mmecomprehensiveevaluationbenchmark}, VQAv2~\cite{goyal2017making}, GQA~\cite{hudson2019gqa}, and MMMU~\cite{yue2024mmmumassivemultidisciplinemultimodal} and report the exact match accuracy following LlaVa~\cite{liu2024improved}. For visual generation, we evaluate Orthus's visual generation performance on GenEval~\cite{ghosh2024geneval} and HPSv2~\cite{wu2023humanpreferencescorev2} following SD3~\cite{esser2024scaling}.
% % , and CIDEr~\cite{vedantam2015cider} scores on the Karpathy test split of MS-COCO~\cite{lin2014microsoft}.
\\
\textbf{Image $\boldsymbol{\rightarrow}$ Text.}
Table~\ref{tab:mmu} shows that: (i) Compared to Chameleon post-trained with the same dataset, Orthus consistently demonstrates superior performance across all benchmarks. 
Besides, inspecting OCR-related tasks in MME-P, we witness a significant superiority of Orthus over Chameleon (with scores of 70 vs. 45). 
% It is also noteworthy that Orthus shows a significant advantage over Chameleon (90 vs. 45) when it comes to OCR-related tasks in MME-P. 
These results validate the superiority of Orthus's modeling by adopting lossless representations for images. (ii) Orthus outperforms other unified models using a single transformer like LWM and Show-o across all benchmarks, highlighting its efficacy for unified modeling. (iii) Compared to larger unified models using an external diffusion model, such as NExT-GPT-13B, Orthus achieves decent results on the VQAv2 benchmark. It is reasonable to speculate that Orthus's potential for multimodal understanding problems can be further unleashed by scaling up training compute and data. 

\begin{figure*}[t]
  \vspace{-6pt}
  \centering
   \includegraphics[width=0.94\linewidth]{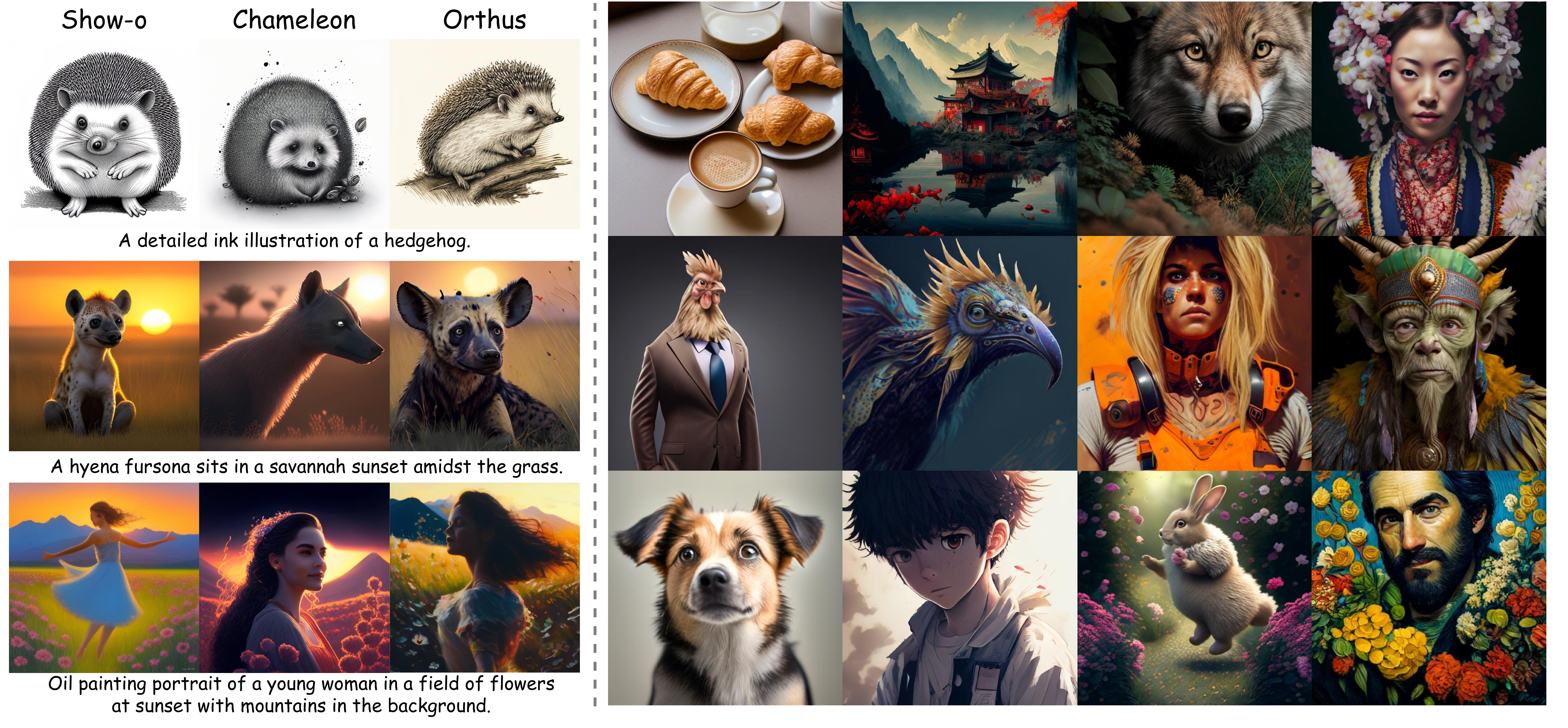}
   \vspace{-5pt}
   \caption{\textbf{Left:} Comparison between images generated by Show-o, Chameleon, and Orthus based on the same prompts. Samples produced by Orthus contain more details. \textbf{Right:} Text-to-image gallery of Orthus.}
   \label{fig:t2i_demo}
   \vspace{-16pt}
\end{figure*}

\textbf{Text $\boldsymbol{\rightarrow}$ image.} \begin{table}[t]
    \centering
    \vspace{-12pt}
    \caption{\textbf{Comparison with state-of-the-arts on visual generation benchmarks.} Model using external pre-trained diffusion model is marked with * and $\text{Chameleon}^\dagger$ is post-trained with the same dataset as Orthus. The results in \textbf{bold} and \underline{underline} are the best and second-best results, respectively.} 
    \vspace{6pt}
    \setlength{\tabcolsep}{2pt}
    \begin{tabular}{clccc}
        \toprule
        \textbf{Type} & \textbf{Model} & \textbf{Res.} & \textbf{GenEval}                                       & \textbf{HPS} \\ 

        \midrule
        \multirow{6}{*}{\shortstack{Gen. \\ Only}} 
        
        % & LlamaGen~\cite{sun2024autoregressive}  & 512 & 0.71 & 0.34 & 0.21 & 0.58 & 0.07 & 0.04 & 0.32 & 26.3 \\
        & SDv1.5~\cite{rombach2022high} & 512 & 0.43 & 27.0 \\
        % & PixArt-$\alpha$~\cite{chen2024pixartalpha} & 512 & 0.98 & 0.50 & 0.44 & 0.80 & 0.08 & 0.07 & 0.48 & - \\
        & SDv2.1~\cite{rombach2022high} & 512 & 0.50 & 27.2 \\
        & DALL-E~\cite{ramesh2022hierarchical} & 512 & 0.52 & 26.9 \\
        & Emu3-Gen~\cite{wang2024emu3} & 512 & 0.54 & - \\
        & SDXL~\cite{podellsdxl} & 512 & 0.55 & 30.9 \\
        & SD3(d=30)~\cite{esser2024scaling} & 512 & 0.64 & - \\
        \midrule
        \multirow{6}{*}{\shortstack{Und. \\ \& \\ Gen.}} 
        & $\text{SEED-X}^{*}$~\cite{ge2024seed} & 448 & 0.49 & - \\
        & LWM~\cite{liu2024world} & 256 & 0.47 & 26.1 \\
        & Show-o~\cite{xie2024show} & 256 & 0.53 & 27.3 \\
        & Transfusion~\cite{zhou2024transfusion} & 256 & \textbf{0.63} & - \\
        & $\text{Chameleon}^{\dagger}$ & 512 & 0.43 & 26.9 \\
        & \textbf{Orthus (Ours)} & 512 & \underline{0.58} & \textbf{28.2} \\
        \bottomrule
    \end{tabular}
    \label{tab:geneval}
    \vspace{-18pt}
\end{table}
Table \ref{tab:geneval} shows that: (i) When compared with strong competitors specialized for text-to-image generations such as DALL-E 2 and SDXL, Orthus achieves an improvement of 0.06 and 0.03 on GenEval, respectively. (ii) Compared to Chameleon and its post-trained version, Orthus demonstrates significant superiority on both GenEval and HPSv2. This advantage can be attributed to the utilization of continuous image representations and diffusion-based continuous modeling, which facilitates the generation of high-quality images with richer detail and stronger alignment with human preferences. (iii) Compared with other unified models such as SEED-X, LWM, and Show-o, Orthus obtains significantly better performance, highlighting the advantages of its modeling strategy. (iv) Qualitative results in Figure \ref{fig:t2i_demo} showcases images generated by Orthus alongside results from other unified models, including Chameleon and Show-o. Results show that Orthus is capable of generating diverse, engaging, and realistic visual imagery at the resolution of 512$\times$512.

\subsection{Ablation Studies}

\textbf{Separate training vs. unified training.} To validate the efficacy of Orthus for unified multimodal modeling, we compare baselines using identical training data but with different learning objectives: (i) a generation-only baseline focused solely on text-to-image generation; (ii) an understanding-only baseline dedicated to visual understanding tasks; and (iii) a unified training objective, the default setting in Orthus. Table \ref{tab:joint_train} shows that both understanding and generation metrics are better through unified training compared to separate task-specific training, highlighting the superiority of Orthus’s modeling which facilitates information gains from bidirectional cross-modal learning.

\begin{table}[t]
    \centering
    \setlength{\tabcolsep}{2.3pt}
    \caption{Comparisons of the performance of Orthus via separate training and unified training across multimodal benchmarks. }
    \vspace{5pt}
    \begin{tabular}{lcccccc}
    \toprule
    Type & $\mathcal{L}_\text{diff}$ & $\mathcal{L}_\text{ar}$ & POPE$\uparrow$ & MME-P$\uparrow$ & GQA$\uparrow$ & GenEval$\uparrow$ \\
    \midrule
    Und. only & \ding{55} & \ding{51} & 78.7 & 1244.2 & 51.9 & - \\
    Gen. only & \ding{51} & \ding{55} & - & - & - & 0.56 \\ 
    Und. \& Gen. & \ding{51} & \ding{51} & \textbf{79.6} & \textbf{1265.8} & \textbf{52.8} & \textbf{0.58} \\
    \bottomrule
    \end{tabular}
    \vspace{-10pt}
    \label{tab:joint_train}
\end{table}

\noindent\textbf{Impact of vision embedding modules on visual understanding tasks.} In this section, we ablate the impact of different choices of vision embedding modules to build Orthus from fully AR models on visual understanding. When we retain the original embedding module in fully AR models (``argmin'' in Table \ref{tab:ablate_t}), a performance drop is observed due to the information loss. Moreover, replacing the embedding module with a randomly initialized linear layer also leads to suboptimal performance due to the significant distribution shift between the embedded space and the transformer’s input space. This misalignment may necessitate training with more image-text pairs to mitigate.

\noindent\textbf{Loss design.} To test the necessity of diffusion modeling for the image features, we train the MLP head with straightforward Mean Squared Error (MSE) loss between predictions and target features. As shown in Appendix \ref{apped:mse}, the model trained with MSE loss generates degraded samples that lack details and exhibit limited color diversity. 
The reason is that the deterministic nature of MSE loss leads to mode collapse. 
% Inspired by Emu-2~\cite{sun2023emu}, which fine-tune large language models (LLMs) using Mean Squared Error (MSE) to reconstruct continuous features from a pre-trained vision encoder and decoding with an external diffusion model accepts continuous feature as condition,

\begin{table}[t]
    \centering
    \caption{Ablation study on the choice of vision embedding modules on visual understanding tasks.}
    \vspace{5pt}
    \setlength{\tabcolsep}{2.5pt}
    \begin{tabular}{lccccc}
    \toprule
    Type & POPE$\uparrow$ & MME-P$\uparrow$ & VQAv2$\uparrow$ & GQA$\uparrow$ & MMMU$\uparrow$ \\
    \midrule
    softmax & \textbf{78.7} & \textbf{1244.2} & \textbf{60.8} & \textbf{51.9} & \textbf{28.0} \\
    argmin & 77.6 & 1064.8 & 57.9 & 50.1 & 26.7 \\ 
    linear & 70.4 & 800.7 & 50.3 & 44.5 & 22.3 \\
    \bottomrule
    \end{tabular}
    \vspace{-10pt}
    \label{tab:ablate_t}
    % \vspace{3.5pt}
\end{table}
% \vspace{-6pt}
\section{Conclusion}
\label{sec:conclusion}
In this paper, we propose Orthus, a unified multimodal model for interleaved image-text understanding and generation. Orthus generates content across modalities by routing the outputs from its shared transformer backbone to modality-specific heads. Its continuous treatment of visual signals preserves input integrity and its unified AR modeling approach for both discrete text tokens and continuous image features enables its superior performance across various multimodal understanding and generation benchmarks. For future work, we plan to scale Orthus by expanding its parameter size and leveraging larger, interleaved datasets to maximize its potential. Furthermore, we aim to broaden its multimodal capabilities by incorporating additional modalities, including video and audio.
% by leveraging efficient inference mechanisms from LLMs to accelerate the AR generation process.

\section*{Impact Statement}
This work presents a challenge in machine learning and proposes a solution, the potential negative consequences are not apparent. While it is theoretically possible for any technique to be misused, the likelihood of such misuse occurring at the current stage is low.
\bibliography{main}
\bibliographystyle{icml2025}

%%%%%%%%%%%%%%%%%%%%%%%%%%%%%%%%%%%%%%%%%%%%%%%%%%%%%%%%%%%%%%%%%%%%%%%%%%%%%%%
%%%%%%%%%%%%%%%%%%%%%%%%%%%%%%%%%%%%%%%%%%%%%%%%%%%%%%%%%%%%%%%%%%%%%%%%%%%%%%%
% APPENDIX
%%%%%%%%%%%%%%%%%%%%%%%%%%%%%%%%%%%%%%%%%%%%%%%%%%%%%%%%%%%%%%%%%%%%%%%%%%%%%%%
%%%%%%%%%%%%%%%%%%%%%%%%%%%%%%%%%%%%%%%%%%%%%%%%%%%%%%%%%%%%%%%%%%%%%%%%%%%%%%%
\newpage
\appendix
\onecolumn
\section{Comparison of Vision Autoencoder}
To construct a vision autoencoder capable of decoding high-quality images based on continuous image features $V$, we freeze the encoder of Chameleon's VQ-VAE, drop the quantization step, and finetune the decoder only to reconstruct images, transforming it into a conventional continuous autoencoder effectively. The decoder is trained on LAION-Aesthetic dataset~\cite{laion-coco}using a learning rate of 1e-5, a batch size of 256, and a total of 15,000 training steps. Table \ref{tab:vae} shows that our vision autoencoder achieves better reconstruction quality compared to the original VQ-VAE. The evaluation is conducted on a subset of the LAION-Aesthetic, consisting of 10,000 images that are excluded from the training dataset.
\label{apped:aoe}
\begin{table}[h]
    \centering
    \vspace{-8pt}
    \caption{Comparison of reconstruction quality for vision autoencoders: the discrete one is worse than the continuous variant.}
    \vspace{6pt}
    \setlength{\tabcolsep}{10pt}
    \begin{tabular}{lcc}
    \toprule
    \textbf{Model}  & \textbf{PSNR}$\uparrow$ & \textbf{SSIM}~\cite{wang2004image}$\uparrow$  \\
    \midrule
    VQ-VAE~\cite{team2024chameleon} & 23.7 & 0.80 \\
    Ours & 26.1 & 0.84 \\
    \bottomrule
    \end{tabular}
    \vspace{-14pt}
    \label{tab:vae}
\end{table}

\section{Training details}
\label{apped:train_detail}
The images for training Orthus-base are the first 10k from \citet{laion-coco}. Both training and evaluation are carried out on servers equipped with 8 NVIDIA A100 80GB GPUs.
 \begin{table}[h]
    \centering
    \vspace{-8pt}
    \caption{Training details for constructing Orthus-base and the instruction-tuned one for visual understanding and generation in \ref{sec:mmu}.}
    \vspace{8pt}
    \setlength{\tabcolsep}{18pt}
    \begin{tabular}{lcc}
    \toprule
     \textbf{Model}  & \textbf{Orthus-base} & \textbf{Instruct-tuning}  \\
    \midrule
    Optimizer & \multicolumn{2}{c}{AdamW ($\beta_1=0.9$, $\beta_2=0.99$)}\\
    Learning Rate & 1e-4 & 1e-5 \\
     Batch Size & 32 & 16 \\
     Training Steps & 15,000 & 35,000 \\
    % Training Hours & 72 & 600 \\
    \bottomrule
    \end{tabular}
    \vspace{-8pt}
    \label{tab:train_details}
\end{table}

% \section{Detailed Scores of HPSv2}
% Table ~\ref{tab:hps} reports scores of HPSv2~\cite{wu2023humanpreferencescorev2} across four categories: `Anime', `Concept art', `Paintings', and `Photo'.
% \label{apped:hps}
% \input{tabs/hps}

\section{Diffusion Loss v.s. MSE Loss}
\label{apped:mse}
\begin{figure}[h]
\centering
\vspace{-6pt}
\includegraphics[width=0.46\textwidth]{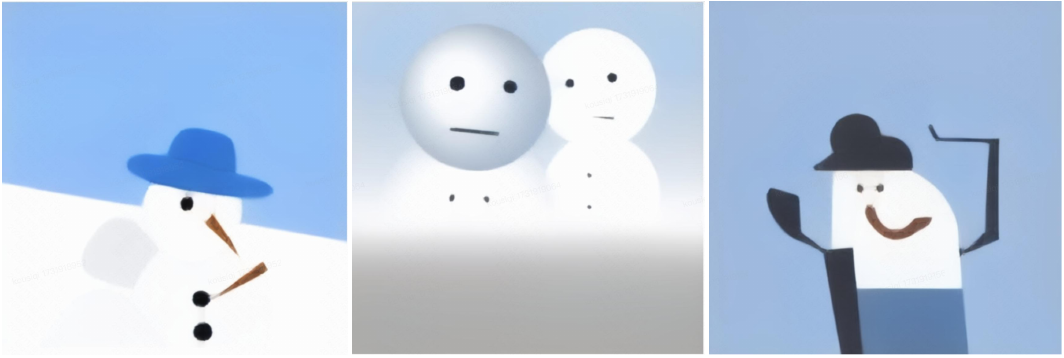} 
\caption{Text-to-image results from models trained with MSE loss. The text prompt is ``Generate an image of a snowman.''}
\vspace{-8pt}
\label{fig:l2}
\end{figure}

\section{Examples on Visual Generation}
\begin{figure*}
    \centering
    \includegraphics[width=0.9\linewidth]{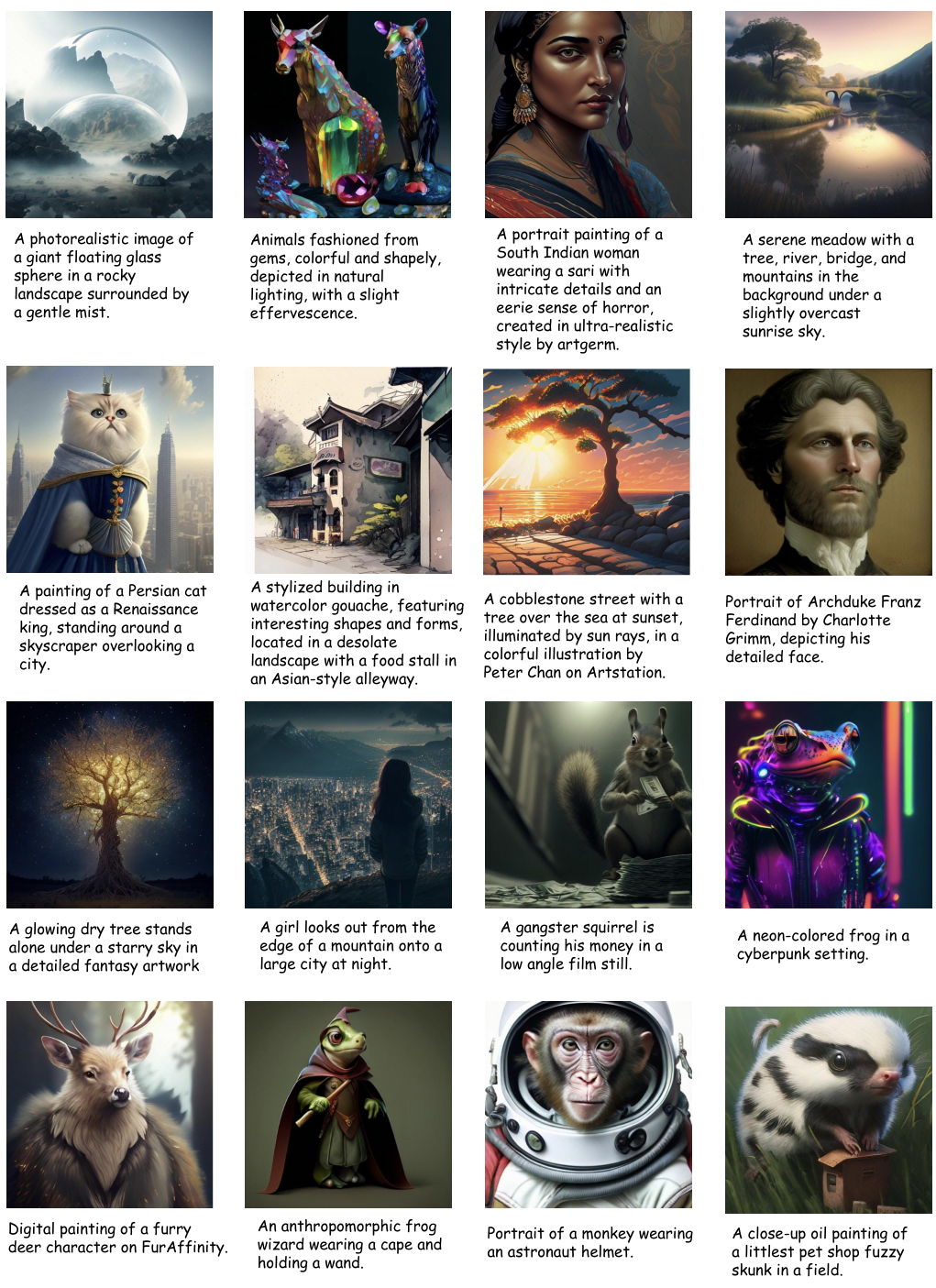}
    \caption{Generated 512 $\times$ 512 images from Orthus. Results demonstrate its ability to generate diverse, engaging, and realistic images.}
    \label{fig:app_t2i}
\end{figure*}

Figure~\ref{fig:app_t2i} shows examples of images generated from Orthus post-trained in Section~\ref{sec:mmu}.

\section{Examples on Visual Understanding}
\begin{figure}[t]
  \centering
\includegraphics[width=0.9\linewidth]{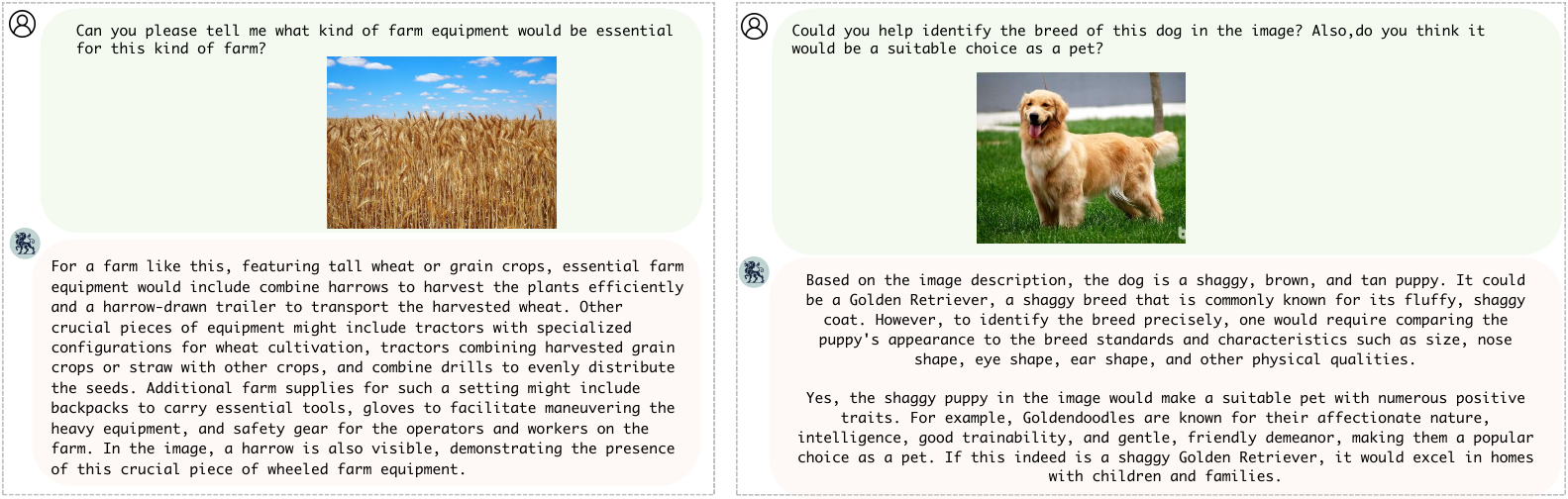}
\caption{Examples of Orthus on visual understanding. Results demonstrate that Orthus exhibits strong instruction-following capabilities and robust generalization abilities.}
\label{fig:mmu}
\end{figure}

In addition to quantitatively evaluating Orthus in Section~\ref{sec:mmu} on domain-specific tasks, we also assess its performance in general chat scenarios in Figure~\ref{fig:mmu}. 

\section{Examples on Image Editing}
\label{apped:edit}
\begin{figure}[t]
  \centering
\includegraphics[width=0.9\linewidth]{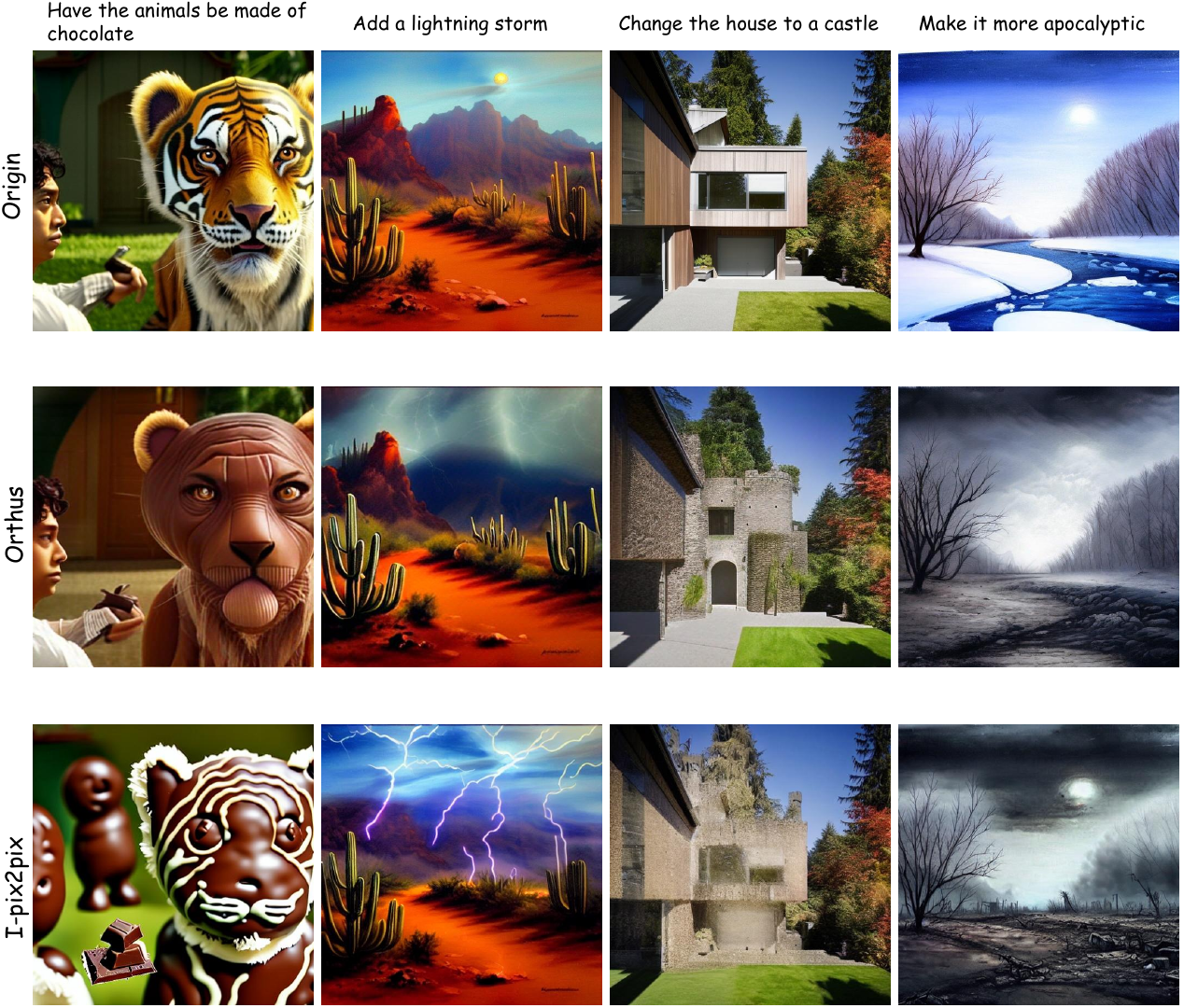}
\caption{Examples of Orthus on image editing. 
Compared to editing-specific diffusion models~\cite{brooks2023instructpix2pix}, Orthus demonstrates better fidelity to the original image in regions where no editing is required.}
\label{fig:app_edit}
\end{figure}

Figure~\ref{fig:app_edit} shows random examples of image editing by Orthus-base post-trained on Instruct-Pix2Pix~\cite{brooks2023instructpix2pix}.

% \begin{figure}[h]
% \centering
% \includegraphics[width=0.45\textwidth]{figs/mmu.pdf} 
% \caption{Qualitative analysis of Orthus on visual understanding.}
% \label{fig:mmu}
% \end{figure}

% \section{Impact of cfg scale}
% \label{sec:rationale}
% % 
% Having the supplementary compiled together with the main paper means that:
% % 
% \begin{itemize}
% \item The supplementary can back-reference sections of the main paper, for example, we can refer to \cref{sec:intro};
% \item The main paper can forward reference sub-sections within the supplementary explicitly (e.g. referring to a particular experiment); 
% \item When submitted to arXiv, the supplementary will already included at the end of the paper.
% \end{itemize}
% % 
% To split the supplementary pages from the main paper, you can use \href{https://support.apple.com/en-ca/guide/preview/prvw11793/mac#:~:text=Delete%20a%20page%20from%20a,or%20choose%20Edit%20%3E%20Delete).}{Preview (on macOS)}, \href{https://www.adobe.com/acrobat/how-to/delete-pages-from-pdf.html#:~:text=Choose%20%E2%80%9CTools%E2%80%9D%20%3E%20%E2%80%9COrganize,or%20pages%20from%20the%20file.}{Adobe Acrobat} (on all OSs), as well as \href{https://superuser.com/questions/517986/is-it-possible-to-delete-some-pages-of-a-pdf-document}{command line tools}.

\end{document}